\documentclass{ecai}
\usepackage{times}
\usepackage{graphicx}
\usepackage{latexsym}
\usepackage{hyperref}
\usepackage{subcaption}
\usepackage{color}
\usepackage{booktabs}
\usepackage{dsfont}

\usepackage{amsthm}
\usepackage{amsmath}

\theoremstyle{definition}
\newtheorem{definition}{Definition}[section]

\ecaisubmission   % inserts page numbers. Use only for submission of paper.
\begin{document}

% \title{Combining Algorithm Selection and Configuration: \\
% A Proof of Concept for the Modular CMA-ES}
\title{Sequential vs. Integrated Algorithm Selection and Configuration:
A Case Study for the Modular CMA-ES}

\author{Diederick Vermetten\institute{Leiden University, The Netherlands} \and Hao Wang$^1$%\institute{Leiden University, The Netherlands, email: } 
 \and Carola Doerr\institute{Sorbonne Universit\'e, CNRS, LIP6, Paris, France} 
 \and Thomas B\"ack$^1$
 %\institute{Leiden University, The Netherlands, email: }
 }

\maketitle
\bibliographystyle{ecai}

\begin{abstract}
    When faced with a specific optimization problem, choosing which algorithm to use is always a tough task. Not only is there a vast variety of algorithms to select from, but these algorithms often are controlled by many hyperparameters, which need to be tuned in order to achieve the best performance possible. Usually, this problem is separated into two parts: \textit{algorithm selection} and \textit{algorithm configuration}. With the significant advances made in Machine Learning, however, these problems can be integrated into a combined algorithm selection and hyperparameter optimization task, commonly known as the CASH problem.  
   
    In this work we compare sequential and integrated algorithm selection and configuration approaches for the case of selecting and tuning the best out of $4,\hspace{-1pt}608$ variants of the Covariance Matrix Adaptation Evolution Strategy (CMA-ES) tested on the Black Box Optimization Benchmark (BBOB) suite. We first show that the ranking of the modular CMA-ES variants depends to a large extent on the quality of the hyperparameters. This implies that even a sequential approach based on complete enumeration of the algorithm space will likely result in sub-optimal solutions. In fact, we show that the integrated approach manages to provide competitive results at a much smaller computational cost.
    
    We also compare two different mixed-integer algorithm configuration techniques,
    called irace and Mixed-Integer Parallel Efficient Global Optimization (MIP-EGO). While we show that the two methods differ significantly in their treatment of the exploration-exploitation balance, their overall performances are very similar. 
\end{abstract}
\section{Introduction}\label{sec:intro}

In computer science, optimization has become an important field of study over the past decades. Because of its rising popularity and its high practical relevance, many different techniques have been introduced to solve particular types of optimization problems. As these methods are developed further, small modifications might lead the algorithm to behave better on specific problem types. However, it has long been know that no single algorithm variant can outperform all others on all functions, as stated in the \textit{no-free-lunch theorem}~\cite{NFL}. 
This fact leads a new set of challenges for practitioners and researchers alike: How to choose which algorithm to use for which problem? 

% \carola{new paragraph here?} 
Even when limiting the scope to a small class of algorithms, the choice of which variant to choose can be daunting, leading practitioners to resort to a few standard versions of the algorithms, which might not be particularly well suited to their problem.
The problem of selecting an algorithm (variant) from a large set is commonly referred to as the \textit{algorithm selection} problem~\cite{kerschke2018survey}. However, the algorithm variant is not the only factor having an impact on performance. The setting of the variable hyperparameters can also play a very important role~\cite{LoboLM07,BelkhirDSS17}. The problem of choosing the right hyperparameter setting for a specific algorithm is commonly referred to as the \textit{algorithm configuration} problem~\cite{EggenspergerLH19}. 

Naturally, the \textit{algorithm selection} and \textit{algorithm configuration} problems are highly interlinked. Because of this, it is natural to attempt to tackle both problems at the same time. Such an approach is commonly referred to as the \emph{Combined Algorithm Selection and Hyperparameter optimization} (CASH) task, which was introduced in~\cite{thornton2013autoweka} and later studied in~\cite{feurer2015efficient,combined_sel_conf,kotthoff2017autoweka}. Note, though, that the by far predominant approach in real-world algorithm selection and configuration is still a sequential approach, in which the user first selects one or more algorithms (typically based on previous experience) and then tunes their parameters (either manually or using one of the many existing software frameworks, such as~\cite{BOHB,SMAC,li2016hyperband,irace,SPOT}), %removed SPOT
before deciding which algorithmic variant to use for their problem at hand. In fact, we observe that the tuning step is often neglected, and standard solvers are run with some default configurations which have been suggested in the research literature (or happen to be the defaults in the implementation). Efficiently solving the CASH problem is therefore far from easy, and far from being general practice. %In the field of machine learning, many different approaches have been used to solve this type of problem~\cite{thornton2013autoweka, kotthoff2017autoweka}. 
%More recently, this type of problem has also been successfully tackled using efficient global optimization~\cite{combined_sel_conf}. % \carola{I would introduce the CASH problem here, before mentioning relevant related work} This has been done quite successfully in~\cite{combined_sel_conf}. 

% \carola{New paragraph here, since this is about our contribution}
In this work, we address the CASH problem in the context of selecting and configuring  variants of the Covariance Matrix Adaptation Evolution Strategy (CMA-ES). The CMA-ES family~\cite{hansen_adapting_1996} is an important collection of heuristic optimization techniques for numeric optimization. In a nutshell, the CMA-ES is an iterative search procedure, which updates after each iteration the covariance matrix of the multivariate normal distribution that is used to generate the samples during the search, effectively learning second-order information about the objective function. Important contributions to the class of CMA-ES have been made over the years, which all reveal different strengths and weaknesses in different optimization contexts~\cite{back_contemporary_2013,hansen_benchmarking_2009}.  
%Also, I would now introduce the modular CMA-ES and then explain what we want to do} Empirically, it has been shown that no single one of these variants outperforms all others~\cite{hansen_benchmarking_2009, back_contemporary_2013}. This paper aims to tune the hyperparameters for these variants, and find the best CMA-ES variant with corresponding hyperparameter settings for given optimization problems.  %\carola{explain CMA-ES and give reference to Hansen paper --- AI researchers may not be familiar with it}. \carola{revise next sentence to. I'd first say that important contributions have been made to this algorithm, which all reveal strength and weaknesses in different optimization contexts. Then introduce modular CMA-ES, explain what has been done already (SSCI paper of Sander) and say that we study here the impact of the hyper-parameter tuning} 
%For CMA-ES, it has been empirically shown that no single variant outperforms all others~\cite{hansen_benchmarking_2009, back_contemporary_2013}. To create these CMA-ES variants, we use 
% \carola{I have revised this part, and merged some parts from Section 2.1:}
%\cite{van_rijn_evolving_2016} suggested a framework which modularizes many popular CMA-ES modifications such that they can be combined to create a total number of $4,\hspace{-1pt}608$ different CMA-ES variants. This modular CMA-ES framework, which is available at~\cite{modCMA}, has been successfully used to study the impact of the eleven different modules on different optimization problems~\cite{van_rijn_evolving_2016}. 
While most of the suggested modifications have been proposed in isolation, \cite{van_rijn_evolving_2016} suggested a framework which modularizes eleven popular CMA-ES modifications such that they can be combined to create a total number of $4,\hspace{-1pt}608$ different CMA-ES variants. It was shown in~\cite{van_rijn_algorithm_2017} that some of the so-created CMA-ES variants improve significantly over commonly used CMA-ES algorithms. This modular CMA-ES framework, which is available at~\cite{modCMA}, provides a convenient way to study the impact of the different modules on different optimization problems~\cite{van_rijn_evolving_2016}. 

The modularity of this framework allows us to integrate the \textit{algorithm selection} and \textit{configuration} into a single mixed-integer search space, where we can optimize both the algorithm variant and the corresponding hyperparameters at the same time. We show that such an integrated approach is competitive with sequential approaches based on complete enumeration of the algorithm space, while requiring significantly less computational effort. We also investigate the differences between two algorithm configuration tools, irace~\cite{irace} and MIP-EGO~\cite{MIP-EGO} (see Section~\ref{sec:hyperparameter} for short descriptions). While the overall performance of these two approaches is comparable, the balance between algorithm selection and algorithm configuration shows significant differences, with irace focusing much more on the configuration task, and evaluating only few different CMA-ES variants. MIP-EGO, in turn, shows a broader exploration behavior, at the cost of less accurate performance estimates. 
% \carola{We need to add here a summary of the key findings!}

% The remainder of this paper is structured as follows: In Section~\ref{sec:related} we introduce the previous work which relates to our topic: the modEA framework, our test-bed and the used performance methods. This section also introduces hyperparameter tuning and explains the methods we use in our experiments. Section~\ref{sec:sequential} introduces a performance baseline by sequential execution of algorithm selection and configuration. We then challenge this baseline in Section~\ref{sec:integrated} by creating a single approach for tuning algorithm variant and hyperparameters simultaneously. Section~\ref{sec:conclusion} concludes the results and future work is discussed in Section~\ref{sec:future}.
\section{Experimental Setup}\label{sec:related}
% \carola{Carola: Related work should be mentioned in the introduction, not here, I have renamed the section}
We summarize the algorithmic framework, the benchmark suite, the performance measures, and the two configuration tools, irace and MIP-EGO, which we employ for the tuning of the hyperparameters.

\subsection{The Modular CMA-ES} 
% \carola{say first (if not already done in the intro) what the CMA-ES is, then motivate the modular CMA-ES framework}
% \carola{I would significantly shorten this section, and just write the following:}  
Table~\ref{tab:es-opt} summarizes the eleven modules of the modular CMA-ES from~\cite{van_rijn_evolving_2016}. Out of these, nine modules are binary and two are ternary, allowing for the total number of $4,\hspace{-1pt}608$ different possible CMA-ES variants.
% , which we will from now on refer to as \textbf{configurations}.\carola{Hm, I still do not like the term ``configuration'' as it is really confusing given that it corresponds to the algorithm selection problem. I anyway guess that a main critique will be that our algorithm selection problem is ``just'' another configuration problem, since it is a structured search space. We should not ``feed'' such feelings, in my opinion.} Each configuration has a number of hyperparameters that need to be set before execution. 

%\carola{1911: I did minor edits here:} 
So far, all studies on the modular CMA-ES framework have used default hyperparameter values~\cite{van_rijn_ppns_2018_adpative,van_rijn_evolving_2016,  research_project}. However, it has been shown that substantial performance gains are possible by tuning these hyperparameters~\cite{andersson2015parameter,BelkhirDSS17}, raising the question how much can be gained from combining the tuning of several hyperparameters with the selection of the CMA-ES variant. In accordance with~\cite{BelkhirDSS17}, we focus on only a small subset of these hyperparameters, namely $c_1, c_c$ and $c_\mu$, which control the update of the covariance matrix. It is well known, though, that other hyperparameters, and in particular the population size~\cite{auger_restart_2005} have a significant impact on the performance as well, and might be much more critical to configure as the ones chosen in~\cite{BelkhirDSS17}. However, we will see that the efficiency of the CMA-ES variants is nevertheless strongly influenced by these three hyperparameters. In fact, we show that the ranking of the algorithm variants with default and tuned hyperparameters can differ significantly, indicating that a sequential execution of \textit{algorithm selection} and \textit{algorithm configuration} will not provide optimal results. 

\begin{table}
\small
    \renewcommand{\arraystretch}{1.1}
    \setlength\tabcolsep{5pt}

    \begin{tabular}{@{}lllll@{}}
    \toprule
    \# & Module name             & 0     & 1                 & 2 \\
    \midrule
    1  & Active Update~\cite{jastrebski_improving_2006}           & off              & on                & - \\
    2  & Elitism                 & ($\mu, \lambda$) & ($\mu{+}\lambda$) & - \\
    3  & Mirrored Sampling~\cite{brockhoff_mirrored_2010}     & off              & on                & - \\
    4  & Orthogonal Sampling~\cite{wang_mirrored_2014}     & off              & on                & - \\
    5  & Sequential Selection~\cite{brockhoff_mirrored_2010}    & off              & on                & - \\
    6  & Threshold Convergence~\cite{piad-morffis_evolution_2015}   & off              & on                & - \\
    7  & TPA~\cite{hansen_cma-es_2008}                     & off              & on                & - \\
    8  & Pairwise Selection~\cite{auger_mirrored_2011}      & off              & on                & - \\
    9  & Recombination Weights~\cite{auger2005quasi_random} & $r_i(\mu)$ & $\frac{1}{\mu}$& - \\
    10 & Quasi-Gaussian Sampling & off             & Sobol            & Halton \\
    11 & Increasing Population~\cite{auger_restart_2005,hansen_benchmarking_2009}   & off             & IPOP             & BIPOP \\
    \bottomrule
    \end{tabular}
        \footnotesize
    \caption{Overview of the CMA-ES modules available in the used framework. 
    The entries in row 9 %, recombination weights, 
    specify the formula for calculating each weight $w_i$, with $r_i:=\log(\mu{+}\frac{1}{2}){-}\frac{\log(i)}{\sum_j w_j}$.}    
    \label{tab:es-opt}
    \vspace{-10pt}
\end{table}

% \begin{table}
%     \centering
%     \begin{tabular}{@{} lr @{}}
%         \toprule
%         Variant                     & Representation\\ %& Short name \\
%         \midrule
%         CMA-ES                      & 00000000000\\% & $CM_0$   \\
%         Active CMA-ES               & 10000000000\\% & $CM_1$   \\
%         Elitist CMA-ES              & 01000000000\\% & $CM_2$   \\
%         Mirrored-pairwise CMA-ES    & 00100001000\\% & $CM_3$   \\
%         IPOP-CMA-ES                 & 00000000001\\% & $CM_4$   \\
%         Active IPOP-CMA-ES          & 10000000001\\% & $CM_5$   \\
%         Elitist Active IPOP-CMA-ES  & 11000000001\\% & $CM_6$   \\
%         BIPOP-CMA-ES                & 00000000002\\% & $CM_7$   \\
%         Active BIPOP-CMA-ES         & 10000000002\\% & $CM_8$   \\
%         Elitist Active BIPOP-CMA-ES & 11000000002\\% & $CM_9$   \\
%         \bottomrule
%     \end{tabular}
%     \caption{Common CMA-ES Variants. A selection of ten common CMA-ES variants is listed here, taken from~\cite{van_rijn_evolving_2016}.}
%         \label{tab:common}
%         \vspace{-20pt}
% \end{table}

\subsection{Test-bed: the BBOB Framework}\label{sec:bbob}
For analyzing the impact of the hyperparameter tuning, we use the Black-Box Optimization Benchmark (BBOB) suite~\cite{hansen_coco:_2016}, which is a standard environment to test black-box optimization techniques. This testbed contains 24 functions $f:\mathds{R}^d\rightarrow\mathds{R}$, of which we use the five-dimensional versions. Each function can be transformed in objective and variable space, resulting in separate instances with similar fitness landscapes. A large part of our analysis is built on data from~\cite{research_project}, which uses the first 5 instances of all functions, for which 5 independent runs were performed on each instance, for each algorithm variant, and each function. This data is available at~\cite{data}. 

\subsection{Performance Measures}
% \carola{Can probably be shortened if we need space --- do we need AHT at all later on?}
% \died{The only time we use AHT is in Figure~\ref{fig:results_full} and the text of Section~\ref{sec:pred_err}, so it can be removed if those parts are rewritten slightly}
% \carola{This whole section can be shortened significantly. Perfectly OK for a Master thesis, but too much for a research paper. I suggest to just keep the first paragraph and to just provide the definitions thereafter, as indicated below:}
%Since we are interested in the performance of the CMA-ES configurations, we need to define exactly which performance measures we are interested in.
We next define the performance measures by which we compare the different algorithms.
First note that CMA-ES is a stochastic optimization algorithm. The number of function evaluations needed to find a solution of a certain quality is therefore a random variable, which we refer to as the \emph{first hitting time}. More precisely, we denote by $t_{i}(v, f, \phi)$ the number of function evaluations that the variant $v$ used in the $i$-th run before it evaluates a solution $x$ satisfying $f(x) \le \phi$ for the first time . %first hitting time of target $\phi$ of run $i$ of variant $v$ on function $f$. 
If target $\phi$ is not hit, we define $t_{i}(v, f, \phi) = \infty$. To be consistent with previous work, such as~\cite{research_project}, we decide to use two estimators of the mean of the hitting time distribution:% average hitting time (AHT) and expected running time (ERT), which are defined as follows.
% In general, there are two distinct approaches which can be taken to analyze the performance of an optimization algorithm. The first is the fixed-budget approach, which looks at the best $f(x)$ value found after a certain budget $B$ has been used. The second approach is fixed-target, which focuses on how many function evaluations are needed by an algorithm to reach a certain target function value $\phi$ for the first time. In this work, we use the fixed-target approach to analyze and compare the configurations created by modEA.

% For the fixed-target analysis, a clear way to define targets across all used benchmark functions needs to be defined. The targets are set as precisions to the optimal values. We use the precision to the optimal value as a shorthand, so $\phi = 10^{-8}$ is hit when $|f_{\text{opt}}-f_{\text{best-so-far}}|\leq \phi$. When we use this shorthand, the hitting time of target $\phi$ is the first function evaluation for which this target is hit. We can write this as $t_{i}(c, f, \phi)$ to signify the hitting time of target $\phi$ of run $i$ of configuration $c$ on function $f$. If target $\phi$ is not hit, we define $t_{i}(c, f, \phi) = \infty$.  

% In essence, the hitting time is an integer-valued random variable, which we denote as $T(c,f, \phi)$. The observed hitting times $t_{i}(c, f, \phi)$ are then sampled from this distribution. To determine the mean of $T$, the most straightforward method is to use the sample mean, also referred to as the Average Hitting Time (AHT), which is defined as follows:
%\carola{1911: I did minor edits here:}
\begin{definition}%[(Penalized) Average Hitting Time]
\label{def:AHT}
For a set of functions $\mathcal{F}=\{f^{(1)}, \ldots, f^{(K)}\}$, the \emph{average hitting time (AHT)} is defined as:
$$\Tilde{T}(v, \mathcal{F}, \phi) = \frac{1}{nK} \sum_{i=1}^n\sum_{j=1}^K \min\{t_{i}(v, f^{(j)}, \phi), P\}$$
When a run does not succeed in hitting target $\phi$, we have $t_{i}(v, f, \phi)=\infty$. In this case, a penalty $P\geq B$ (where $B$ is the maximum budget) is applied. Usually, this penalty is set to $\infty$, in which case this value is called AHT. Otherwise, it is commonly referred to as \emph{penalized} AHT.
%\end{definition}
%\vspace{-3mm}
% The AHT is a very simple estimator for the mean of the hitting time but more often the so-called Expected Running Time (ERT) is used. This is defined as follows:
% \begin{definition}[Expected Running Time]

In contrast, the \emph{expected running time (ERT)} equals 
$$\text{ERT}(v, \mathcal{F}, \phi) = \frac{\sum_{i=1}^n\sum_{j=1}^K \min\{ t_{i}(v, f^{(j)}, \phi), B\}}{\sum_{i=1}^n\sum_{j=1}^K \mathds{1}\{t_{i}(v, f^{(j)}, \phi) <\infty\}}.$$
\end{definition}
Previous work has shown ERT, as opposed to AHT, to be a consistent, unbiased estimator of the mean of the distribution hitting times~\cite{auger_restart_2005}. However, it is good to note that ERT and AHT are equivalent when all runs of variant $v$ manage to hit target $\phi$. 

In the context of the modular CMA-ES, the CASH problem is adopted as follows. Given a set of CMA-ES variants $V$, a common hyperparameter space $H$, a set of function instances $\mathcal{F}$, and a target value $\phi$, the CASH problem aims to find the combined algorithm and hyperparameter setting that solves the problem below:
$$v^*, \mathbf{h}^* = \underset{v\in V, \mathbf{h}\in H}{\mathrm{argmin}}\operatorname{ERT}\left(v_h, \mathcal{F}, \phi\right).$$
Note here that we aim at finding the best (variant,hyperparameter)-pair for each of the 24 BBOB functions individually and we consider as $\mathcal{F}$ the set of the first five instances of each function. We do not aggregate over different functions, since the benchmarks can easily be distinguished by exploratory landscape approaches~\cite{BelkhirDSS17}.
% \died{Does that make sense tough? the ERT is taken over the aggregate of all runs, not the mean of separate ERTs, and the target values are always the same for all function instances}
%To avoid overfitting to the selected random seeds, %we perform the \textit{algorithm selection} and \textit{configuration} on $25$ runs, and 
%we validate the ERT using a large set of $250$ runs.

% \subsection{Overview of terminology}
% To remain consistent with previous work, we use the following terminology:
% \begin{itemize}
%     \item \textbf{Module}: A single option within a configurable algorithm. For example: elitism in modEA.
%     \item \textbf{Configuration}: An instantiation of a configurable algorithm according to a complete set of module settings. For example: the common configurations in Table~\ref{tab:common}.
%     \item \textbf{Target}: A predefined value indicating a precision to the optimal value, which can be reached during an optimization run. A target is hit when $|f_{\text{opt}}-f_{\text{best-so-far}}|\leq \tau$. We use targets between $10^2$ and $10^{-8}$.
%     \item \textbf{Instance}: A specific instantiation of a function. For example: instance 1 of BBOB-function 21. 
%     \item \textbf{Run}: A single execution of a configuration on a single instance of a function.
% \end{itemize}

\subsection{Hyperparameter Tuning}\label{sec:hyperparameter}
% \carola{I would merge the whole first part into Section 2.1 and mention here only: We compare in this work two different off-the-shelf tools for hyperparameter tuning: irace and MIP-EGO.}
% To tune the CMA-ES configurations available in modEA, different hyperparameters can be selected to be optimized. In this work, we decided to follow the approach from~\cite{nacim_thesis}, which focuses on the tuning of three important learning rates within CMA-ES: $c_1, c_c$ and $c_\mu$, which control the update of the covariance matrix. To analyze the robustness of our results, we have compared two different off-the-shelf tools for the tuning of these hyperparameters, irace and MIP-EGO.%\carola{, which determine the speed of XX, XX, and XX, respectively (add intuitive description)}.
%To tune these hyperparameters, two tuning algorithms have been used: MIP-EGO and Irace.

In this work, we compare two different off-the-shelf tools for mixed-integer hyperparameter tuning: irace and MIP-EGO.

% \begin{figure}
%     \centering
%     \includegraphics[width=0.35\textwidth, trim={50 20 0 60},clip]{img/EGO_Flowchart3.pdf}
%     \caption{Basic Flowchart for EGO}
%     \label{fig:EGO_flowchart}
% \end{figure}

%\subsubsection{Irace}\label{sec:irace}

% \begin{figure}
%     \centering
%     \includegraphics[width=0.5\textwidth]{img/irace_illustration.png}
%     \caption{Illustration of basic iterated racing.}
%     \label{fig:iterated_racing}
% \end{figure}

\textbf{Irace}~\cite{irace, irace_2011} is an algorithm designed for hyperparameter optimization, which implements an iterated racing procedure. irace is implemented in R and is freely available at~\cite{irace_code}. For our experiments, we use the elitist version of irace with adaptive capping, which we briefly describe in the following. 

irace works by first sampling a set of candidate parameter settings, which can be any combination of discrete, continuous, categorical, or ordinal variables. These parameters are empirically evaluated on some problem instances, which are randomly selected from the set of available instances. After running on $x$ instances, a statistical test is performed to determine which parameter settings to discard. The remaining parameter settings are then run on more instances and continuously tested every $\ell$ iterations until either only a minimal number of candidates remain or until the budget of the current iteration is exhausted. The surviving candidates with the best average hitting times are selected as the elites.

After the racing procedure, new candidate parameter settings are generated by selecting a parent from the set of elites and ``mutating'' it, as described in detail in~\cite{irace}. After generating the new set of candidates, a new race is started with these new solutions, combined with the elites. Since we use an elitist version of irace, these elites are not discarded until the competing candidates have been evaluated on the same instances which the elites have already seen. This is done to prevent the discarding of candidates which perform well on the previous race based on only a few instances in the current race. 

% \begin{figure}
%     \centering
%     \includegraphics[width=0.5\textwidth]{img/irace_illustration2.png}
%     \caption{Illustration of elitist iterated racing as implemented in the irace-package. Figure taken from~\cite{irace_capping}.}
%     \label{fig:irace_illustration}
% \end{figure}

Apart from using elitism and statistical tests to determine when to discard candidate solutions, we also use another recently developed extension of irace, the so-called \emph{adaptive capping}~\cite{irace_capping} procedure. Adaptive capping helps to reduce the number of evaluations spent on candidates which will not manage to beat the current best. Adaptive capping enables irace to stop evaluating a candidate once it reaches a mean hitting time which is worse than the median of the elites, indicating that this candidate is unlikely to be better than the current best parameter settings.

%\subsubsection{MIP-EGO}\label{sec:mipego}
\textbf{Mixed-Integer Parallel Efficient Global Optimization (MIP-EGO)}~\cite{MIP-EGO, wang2017new} is a variant of Efficient Global Optimization (EGO, a sequential model-based optimization technique), which can deal with mixed-integer search-spaces. Because EGO is designed to deal with expensive function evaluations, and this variant has the ability to deal with continuous, discrete, and categorical parameters, it is also well suited to the hyperparameter tuning task. It uses a much different approach as irace, as we will describe in the following. 

EGO works by initially sampling an set of solution candidates from some specified probability distribution, specifically a Latin hypercube sampling in MIP-EGO. Based on the evaluation of these initial points, a meta-model is constructed. Originally, this was done using Gaussian process regression, but MIP-EGO uses random forests to be able to deal with mixed-integer search spaces. Based on this model, a new point (or a set of points) is proposed according to some metric, called the \emph{acquisition function}. This can be as simple as selecting the point with the largest \textit{probability of improvement} (PI) or the largest \textit{expected improvement} (EI). More recently, acquisition  functions based on moment-generating function of the improvement have also been introduced~\cite{wang2017new}. For this paper, we use the basic EI acquisition function, which is maximized using a simple evolution strategy. After selecting the point(s) to evaluate, the meta-model is updated according to the quality of the solutions. The process is repeated until a termination criterion (budget constraint in our case) is met. 
\section{Baseline: Sequential methods}\label{sec:sequential}
To establish a baseline of achievable performance of tuned CMA-ES variants, we propose a simple sequential approach of algorithm selection and hyperparameter tuning. Since the ERT for all variants on all benchmark functions is available, a complete enumeration technique would be the simplest form of algorithm selection. Then, based on the required robustness of the final solution, either one of several algorithm variants can be selected to undergo hyperparameter tuning. More precisely, we define two sequential methods as follows:
\begin{itemize}
    \item \textbf{Na\"ive sequential}: Perform hyperparameter tuning (using MIP-EGO) on the one CMA-ES variant with the lowest ERT
    \item \textbf{Standard sequential}: Perform hyperparameter tuning (using MIP-EGO) on a set of 30 variants. We have chosen to consider the following set of variant in order to have a wide representation of module settings, and to be able to fairly compare the impact of hyperparameter tuning across functions:%, selected in three groups as follows: \carola{this is not so well-motivated. Why 10 and why 200-210? Maybe just say: we have chosen to consider the following approach or something like this, but if you can motivate the choice a bit, it would help}
    \begin{itemize}
        \item The 10 variants with lowest ERT.
        \item The 10 variants ranked 200-210 according to ERT.
        \item 10 `common' variants, i.e., CMA-ES variants previously studied in the literature (see Table~V in~\cite{van_rijn_evolving_2016}).
        %The 10 'common' configurations from Table~\ref{tab:common}.
    \end{itemize}
\end{itemize}

For both of these methods, the execution of MIP-EGO has a budget of $200$ ERT-evaluations, each of which is based on $25$ runs of the underlying CMA-ES variant (i.e., 5 runs per each of the five instances). Since the observed hitting times show high variance, we validate the ERT values by performing 250 additional runs (50 runs per each instance). All results shown will be ERT from these verification runs, unless stated otherwise. The variant selection and hyperparameter tuning is done separately for each function. 
% \carola{ (we need to specify this, as sometimes people will assume that you are looking for hyperparameters that work well across all functions)} 

\subsection{First Results}
% To judge the performance of the sequential approaches, their final ERT can be compared to the ERT achieved by the default version of CMA-ES, as well as the configuration selected by the algorithm selection, without any hyperparameter tuning. \carola{I would first show a comparison of naive vs standard sequential.} \died{Do you mean something like the figure 'seq\_naive\_vs\_normal\_' variants 1 and 2 (in img\_2 folder)?} \carola{Then compare standard sequential against standard CMA-ES and against best default config per function (chart similar to Fig 4)} \died{Something like variants 3 and 4 of 'seq\_naive\_vs\_normal\_' (compared to default cma-es and best config respectively)?} \carola{, then do the ECDF curve (and remember to spell out what ECDF stands for)  "While Fig X only shows ERT comparisons, ECDF curves allow to investigate the advantage of the tuning across the whole optimization process. We present such an ECDF curve in Fig.~X, where we aggregate over all 24 functions. While the curves may seem to be very close at first sight, note that XXX...} \carola{I would probably remove Fig 4, since this is rather also one of the pitfalls, and since these are not the final results anyway, which will be the ones shown in Fig 8. You can just verbally describe that while XX out of the  functions show an improvement of XX\% on average (arnge xx\%-xx\%), some functions show a negative improvement, which may sound counter-intuitive etc. Maybe move this to Section 3.2.1?} 

While the two sequential methods introduced are quite similar, it is obvious that the na\"ive one will always perform at most equally well as the standard version, since the algorithm variant tuned in the na\"ive approach is always included in the set of variants tuned by the standard method (the same tuned data is used for both methods to exclude impact of randomness). In general, the standard sequential method achieves ERTs which are on average around $20\%$ lower than the na\"ive approach. 

\begin{figure}
    \centering
    \includegraphics[width=0.5\textwidth]{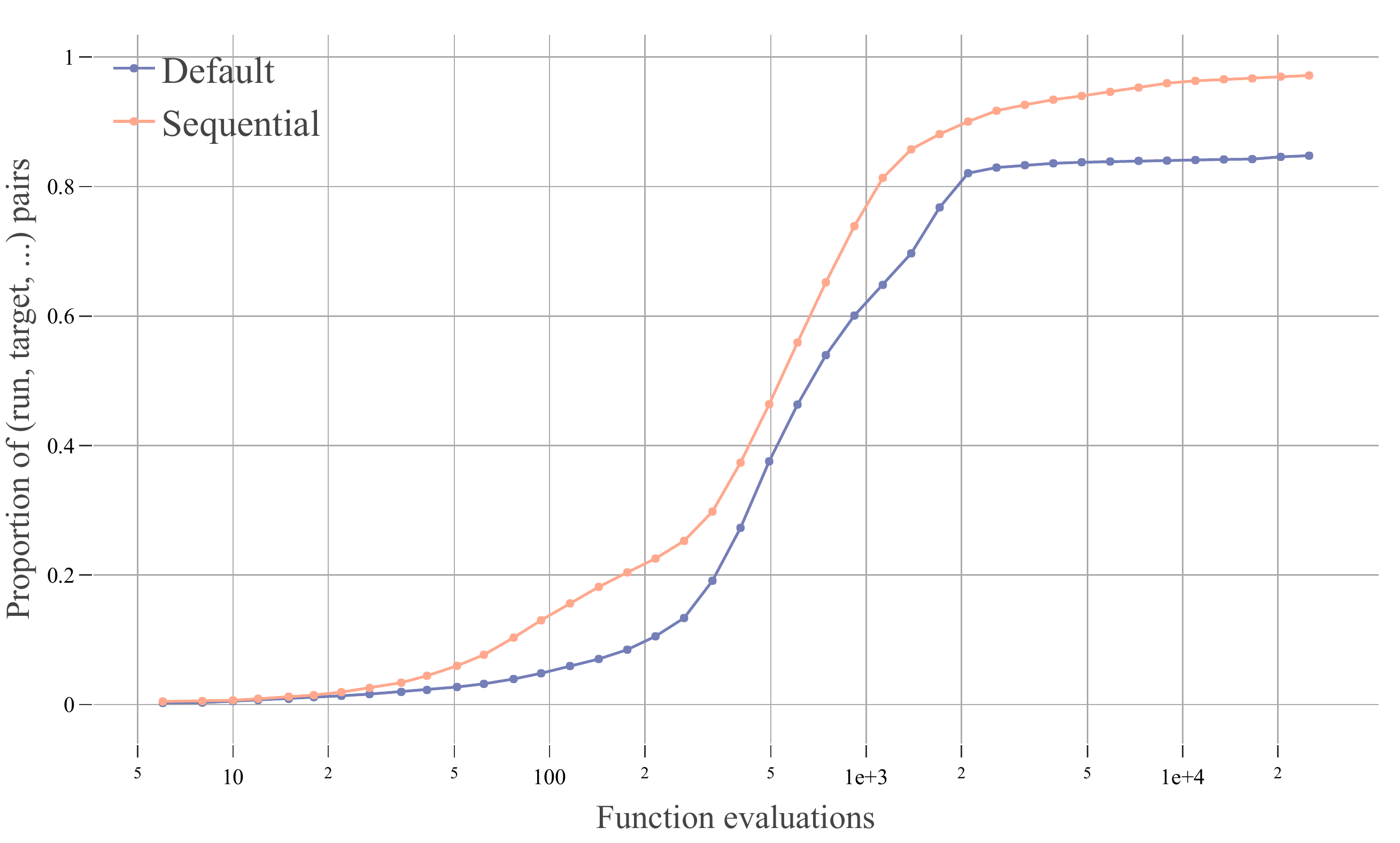}
    \caption{ECDF curve over all benchmark functions of both the \textbf{standard sequential} method as well as the default CMA-ES. Figure generated using IOHprofiler~\cite{IOHprofiler}. %\carola{I would start the x-axis at 10}
    }
    \label{fig:ECDF_seq_def}
\end{figure}

To better judge the quality of these sequential methods, we compare their performance to the default variant of the CMA-ES, which is the variant in which all modules are set to 0. This can be done based on ERT, for each function, but that does not always show the complete picture of the performance. Instead, the differences between the performances of the sequential method and the default CMA-ES are shown in a \emph{Empirical Cumulative Distribution Function (ECDF)}, which aggregates all runs on all functions and shows the fraction of runs and targets which were hit within a certain amount of function evaluations. This is shown in Figure~\ref{fig:ECDF_seq_def} (targets used available at~\cite{data_thesis}). From this, we see that the sequential approach completely dominates the default variant. When considering only the ERT, this improvement is on average $73\%$. 

As well as comparing performance against the default CMA-ES, it can also be compared against the best modular variant with default hyperparameters, i.e. the result of pure algorithm selection. For this, the standard sequential approach manages to achieve a $24.7\%$ improvement in terms of average ERT, as opposed to $6.3\%$ for the na\"ive version. Of note for the na\"ive version is that not all comparisons against the pure algorithms selection are positive, i.e. for some ($5$) functions it achieves a larger ERT. This might seem counter-intuitive, as one would expect hyperparameter tuning to only improve the performance of an algorithm. However, this is where the inherent variance of evolution strategies has a large impact. In short, because ERTs seen by MIP-EGO are based on only $25$ runs, it 
%is quite likely 
may happen that a sub-optimal hyperparameter setting will be selected. This is explained in more detail in the following sections.

% The comparison to the default CMA-ES can be done using the ECDF-plot as shown in Figure~\ref{fig:ECDF_seq_def}. From this figure, it is clear that the standard sequential approach dominates the default CMA-ES. In terms of ERT, the average improvement is $73\%$.

% % \begin{figure}
% %     \centering
% %     \includegraphics[width=0.5\textwidth]{img/Rel_impr_best_def_var2.pdf}
% %     \caption{Relative improvement in ERT over the best configuration with default hyperparameters, both for the \textbf{standard sequential} (denoted by’Best Pair’) as for the \textbf{na\"ive sequential} methods (denoted by ‘Tuned version’).\died{I will update the legend in this figure}}
% %     \label{fig:comp_seq_algsel}
% % \end{figure}

% A better indication of the performance of the sequential methods would be the comparison to the result of algorithm selection, without any hyperparameter tuning. This is visualized in Figure~\ref{fig:comp_seq_algsel}, where it is shown that the average improvement is quite positive. However, there are some negative improvements shown as well. This is counter-intuitive, as one would expect hyperparameter tuning to only improve the performance of an algorithm. However, this is where the inherent variance of evolution strategies has a large impact. In short, because ERTs seen by MIP-EGO are based on only $25$ runs, it is quite likely that a sub-optimal hyperparameter setting will be selected. This is explained in more detail in the following sections.

\subsection{Pitfalls}
The sequential methods described here have the advantage of being based on algorithm selection by complete enumeration. In theory, this would be the perfect way of selecting an algorithm variant. However, since CMA-ES are inherently stochastic, variance has a large effect on the ERT, and thus on the algorithm selection. This might not be an issue if one assumes that hyperparameter tuning has an equal impact on all CMA-ES variants. Unfortunately, this is not the case in practice.

\subsubsection{Curse of High Variance}\label{sec:variance}
The inherent variance present in the ERT-measurements does not only cause potential issues for the algorithm selection, it also plays a large role in the hyperparameter configuration. As previously mentioned, the ERT after running MIP-EGO can be larger than the ERT with the default hyperparameters, even tough the default hyperparameters are always included in the initial solution set explored by MIP-EGO. Since this might seem counter-intuitive, a small-scale experiment can be designed to show this phenomenon in more detail. 
% \carola{If we run out of space, we can consider to remove this experiment, but I suggest to keep it for the time being}
\begin{figure}[!hb]
\vspace{-5mm}
    \centering
    \includegraphics[width=0.5\textwidth, trim=0mm 0mm 0mm 7mm, clip]{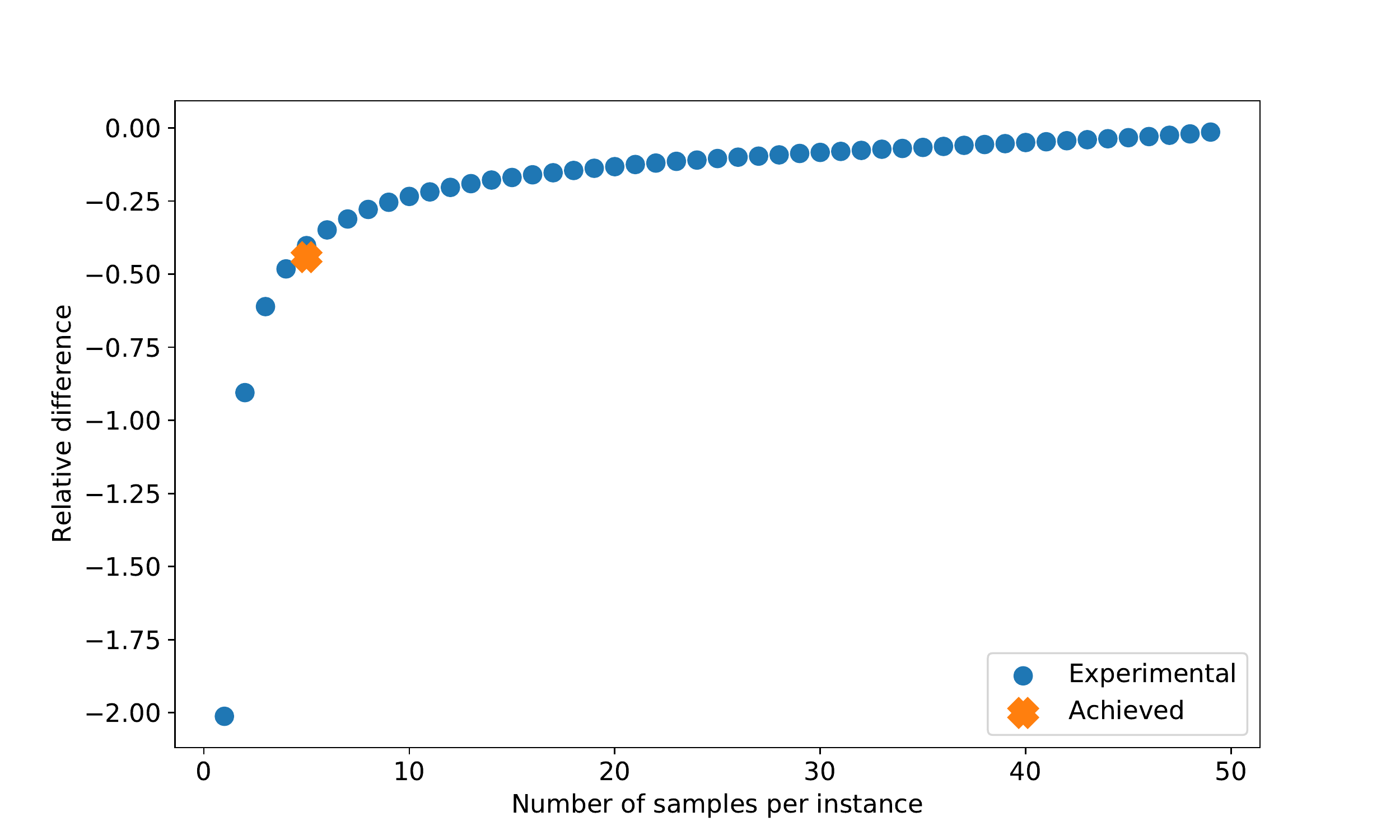}
    \caption{Average  improvement  of  ERT  obtained  from  250  runs  vs  the  value  obtained after running MIP-EGO (25 runs) in orange, vs experimental improvement. Experimental improvement obtained over 100 repetitions of selecting $k$ samples per instance for each variant and calculating their respective ERT.}
    \label{fig:pred_errs_F21}
\end{figure}

This experiment is set up by first taking the set of 50 hitting times for each instance as encountered in the verification runs. Then, sample $x$ runs per instance from these hitting times and calculate the resulting ERT. Repeat this $10$ times, and take the minimal ERT. Then we can compare the original ERT to this new value. This is similar to the internal data seen by MIP-EGO, if we assume that $5\%$ of the variants it evaluated have a similar hitting time distribution. When preforming this experiment on a set of $100$ algorithm variants on F21, we obtain the results as seen in Figure~\ref{fig:pred_errs_F21}, which shows that the actual differences between ERTs given by MIP-EGO and those achieved in the verification runs matches the difference we would expect based on this experiment.

\subsubsection{Differences in Improvement}\label{sec:differences_impr}

% \carola{as suggested in the skype call, I would add here a figure similar to Fig 5 (?) in the thesis}
Even when accounting for the impact of stochasticity on algorithm selection, there can still be large differences in the impact of hyperparameter tuning for different variants. This can be explained intuitively by the notion that some variants have hyperparameters which are already close to optimal for certain problems, while others have very poor hyperparameter settings. Hyperparameter tuning might then lead to some variants, which perform relatively poorly with default hyperparameters, to outperform all others when the hyperparameters have been sufficiently optimized.

\begin{figure}
    \centering
    \includegraphics[width=0.5\textwidth, trim={25 0 30 30},clip]{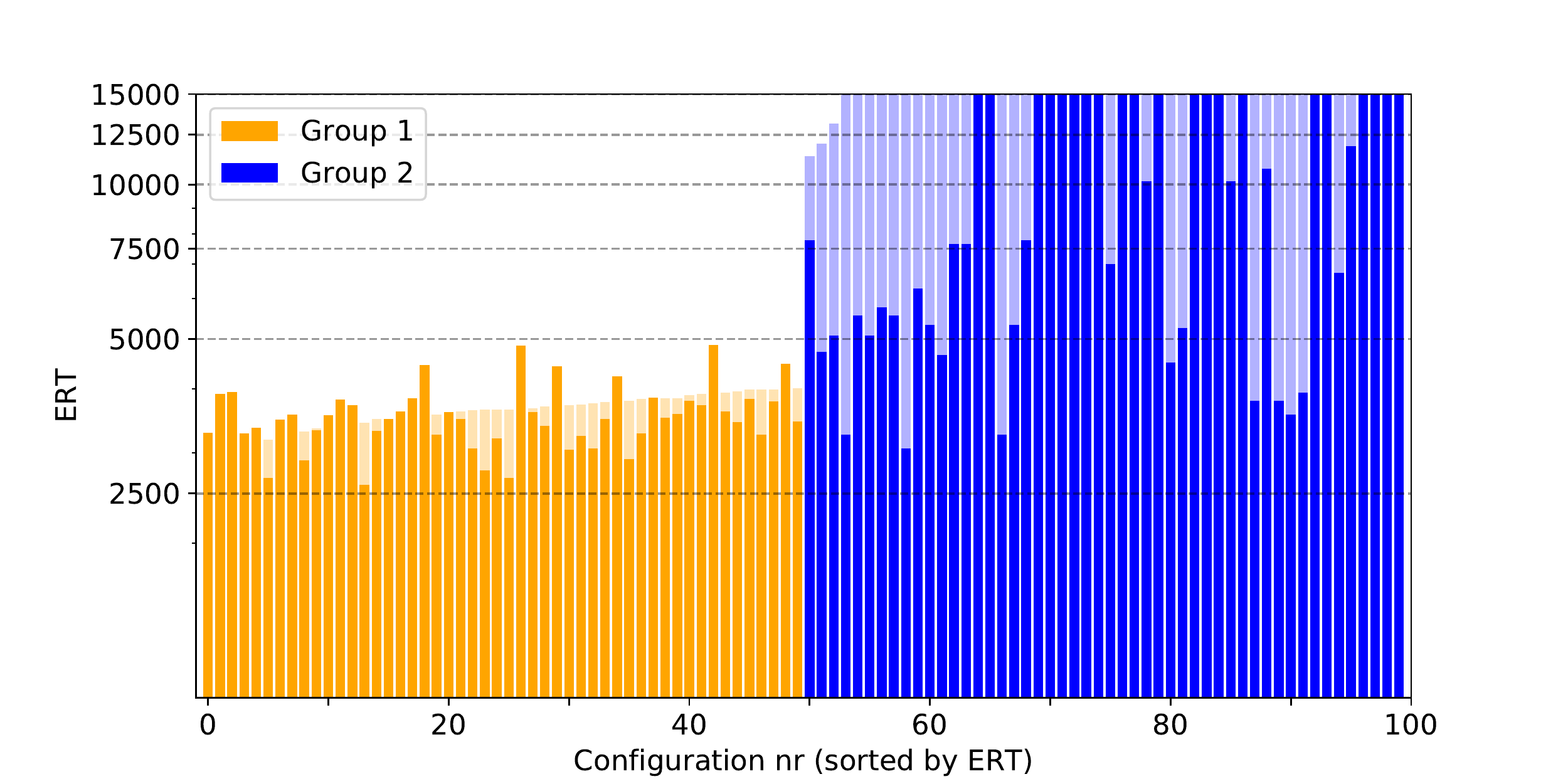}
    \caption{ERT for two groups of CMA-ES variants. The ERT before tuning is shown in light color (based on 25 runs), while the ERT after tuning and verification is shown in a darker shade.}
    \label{fig:ERT_F12}
\end{figure}

This can be shown clearly by looking at one function in detail, F12 in this case, and studying the impact of hyperparameter tuning on two sets of algorithm variants. The two groups are selected as follows: the top 50 according to ERT, and a set of 50 variants. Then, for each of these variants, the hyperparameters are tuned using MIP-EGO. The resulting ERTs are shown in Figure~\ref{fig:ERT_F12}. From this figure, it is clear that the relative improvements are indeed much larger for the group of random variants. There are even some variants which start with very poor ERT, which after tuning become competitive with the variants from the first group. In this first group, the effects noted in Section~\ref{sec:variance} are also clearly present, with some variants performing worse after tuning than before.

\begin{figure}
    \centering
    \includegraphics[width=0.5\textwidth]{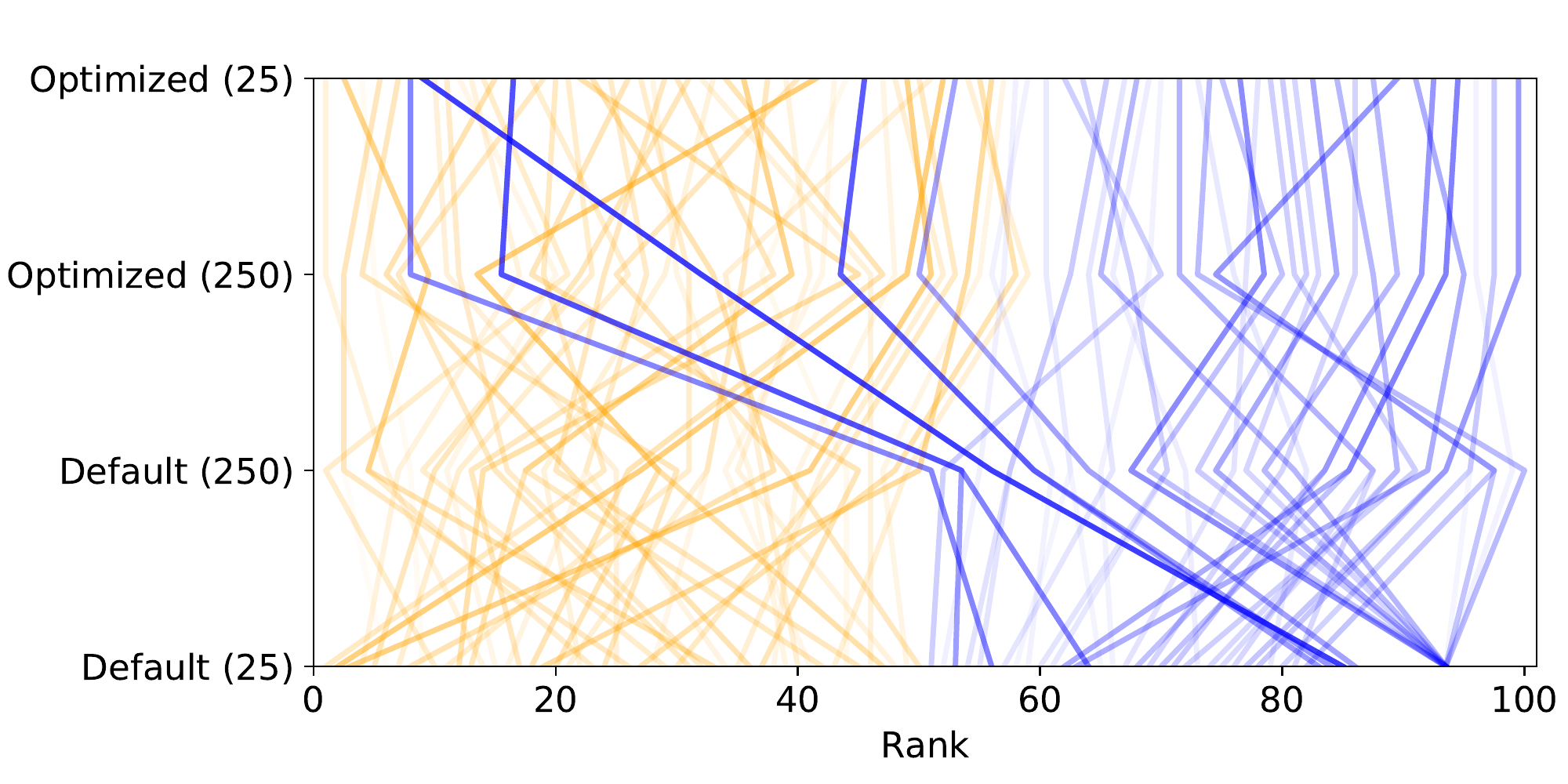}
    \caption{Evolution of ERT-based ranking (lower rank is better) of $100$ algorithm variants on F12. Default refers to the ERT using the default hyperparameters while optimized is the best ERT using the tuned hyperparameters as found by MIP-EGO. Darker lines correspond to larger changes in ranking. Colors correspond to the grouping of variants.}
    \label{fig:ranking_F12}
    \vspace{0pt}
\end{figure}

We can also rank these CMA-ES variants based on their ERT, both with the default and tuned hyperparameters, both for the $25$ runs seen during the tuning as the $250$ verification runs. The resulting differences in ranking are then shown in Figure~\ref{fig:ranking_F12}. This figure shows both the impact of variance on the $25$-run rankings as the much larger differences present between the rankings with default versus tuned hyperparameters.

\begin{figure}
    \centering
    \includegraphics[width=0.5\textwidth, trim={25 0 30 30},clip]{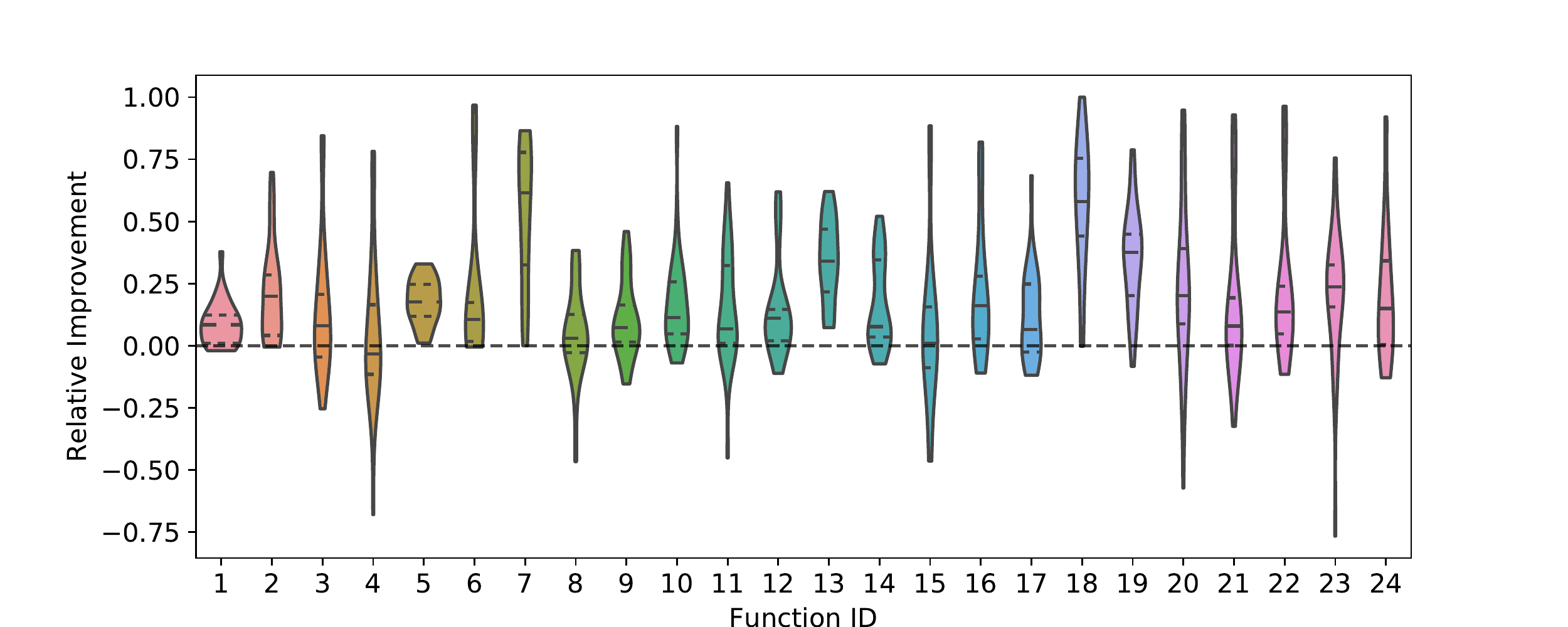}
    \caption{Distribution of relative improvement in ERT between the default and tuned hyperparameters. For each function, 30 variants are tuned with MIP-EGO, and the resulting (variant, hyperparameters)-pairs are rerun 250 times to validate the results. The same is done for the default hyperparameters, and then the relative improvement in ERT is calculated.}
    \label{fig:tuning_per_fid}
    \vspace{-10pt}
\end{figure}

These differences in improvement after hyperparameter tuning are also highly dependent on the underlying test function. When executing the sequential approach mentioned previously, 30 variants are tuned for each function, and the ERTs are verified using $250$ runs. The resulting data can then give some insight into the difference in terms of relative improvement possible per function, as is visualized in Figure~\ref{fig:tuning_per_fid}. This shows that, on average, a relatively large performance improvement is possible for the selected variants. However, the distributions are have large variance, and differ greatly per function. This highlight the previous findings of different variants receiving much greater benefits from tuned hyperparameters than others, thus confirming the results from Figure~\ref{fig:ranking_F12}.

\subsubsection{Scalability}

The final, and most important, issue with the sequential methods lies in their scalability. Because these methods rely on complete enumeration of all variants based on ERT, the required number of function evaluations grows as the algorithm space increases. If just a single new binary module is added, the size of this space doubles. This exponential growth is unsustainable for the sequential methods, especially if the testbed will also be expanded to include higher-dimensional functions (requiring more budget for the runs of the CMA-ES).

\section{Integrated Methods}\label{sec:integrated}
To tackle the issue of scalability, we propose a new way of combining algorithm selection and hyperparameter tuning. This is achieved by viewing the variant as part of the hyperparameter space, which is easily achieved by considering the module activations as hyperparameters. This leads to a mixed-integer search space, which both MIP-EGO and irace can easily adapt to. Thus, we will use two integrated approaches: MIP-EGO and irace. Both will get a total budget of $25,\hspace{-1pt}000$ runs, which irace allocates dynamically while MIP-EGO allocates $25$ runs to calculate ERTs for its solution candidates. 

\subsection{Case Study: F12}

\begin{figure}
    \centering
    \includegraphics[width=0.5\textwidth, trim={25 0 30 40},clip]{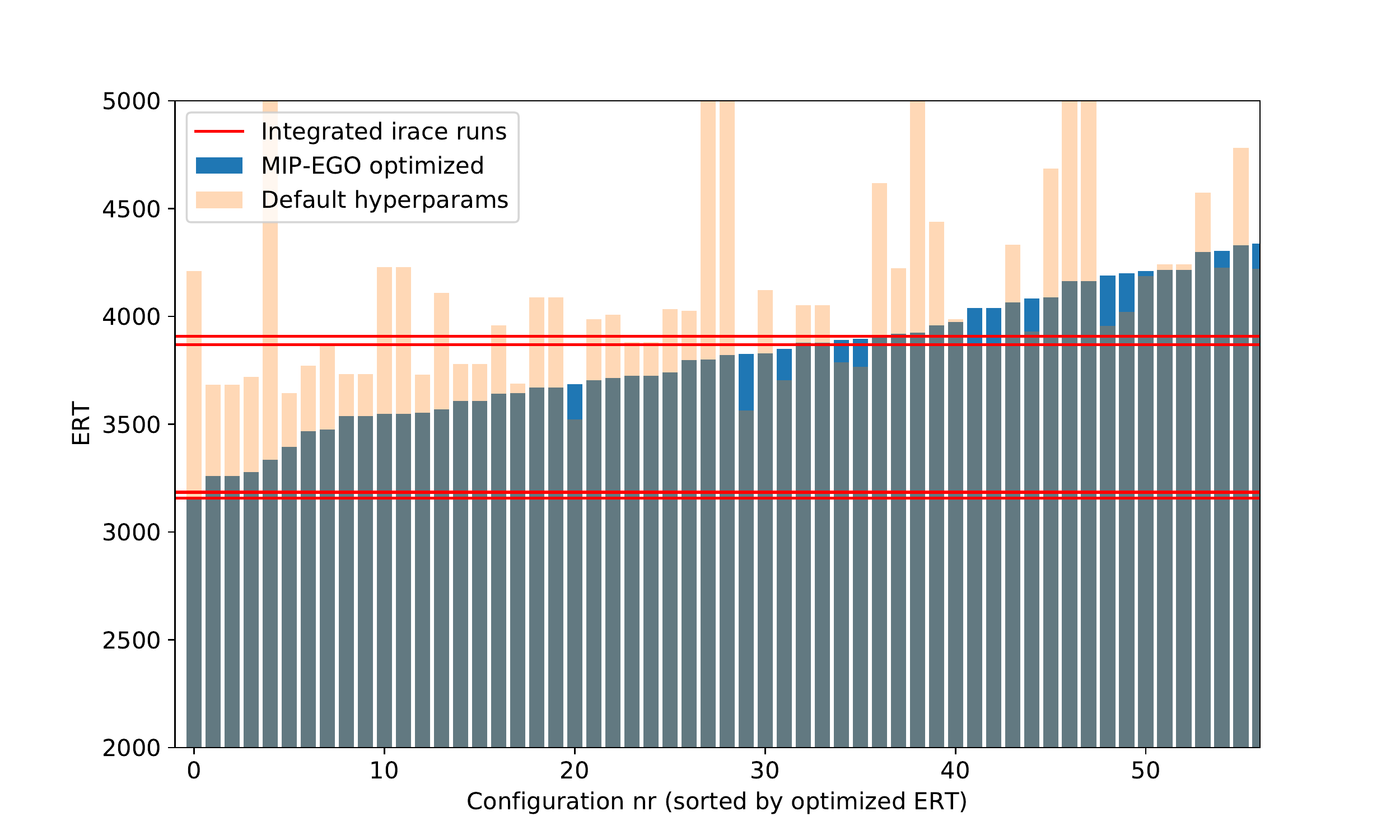}
    \caption{Comparison of ERTs achieved by the integrated approach using irace and the tuning a set of 56 tuned variants (using MIP-EGO).}
    \label{fig:integrated_F12}
\end{figure}

The viability of this integrated approach can be established by looking at a single function and comparing the results from the integrated approach to the previously established baselines. This is done for F12, since for this function, data for the top 50 variants is available, as shown in Figure~\ref{fig:ERT_F12}. We run irace 4 times on instance 1 of this function, and compare the result to those achieved by the best tuned variants. This is done in Figure~\ref{fig:integrated_F12}. From this figure, it can be seen that two of the runs from irace are very competitive with the best tuned variants, while the other two still manage to outperform most variants with default hyperparameters. This shows that this integrated approach is quite promising, and worth to study in more detail. 
% However, this figure also gives rise to another question: What configurations does irace find, and how do they compare to the best configurations with default hyperparameters?

% This question can be answered by looking at the module activations found by irace and comparing this the top 50 configurations, which is visualized in the combined module activation plot in Figure~\ref{fig:cma_F12}. From this figure, it can be seen that the modules for which there is a consensus among the top configurations, which indicate that these modules seem to be important for this function, irace finds the same values. This leads us to conclude that irace finds relatively well performing configurations, with the added benefit of tuning the hyperparameters at the same time.

% \begin{figure}
%     \centering
%     \includegraphics[width=0.5\textwidth]{img_2/F12_CMA_with_irace.pdf}
%     \caption{Combined module activation plot of the configurations found by the integrated irace approach (black) and the top 50 configurations with default hyperparameters (colored). }
%     \label{fig:cma_F12}
% \end{figure}

% \carola{I would include here, early on, the picture with the 4 red horizontal lines for the integrated optimization with irace }

\subsection{Results}
\begin{figure}
    \centering
    \includegraphics[width=0.5\textwidth, trim={50 10 50 35},clip]{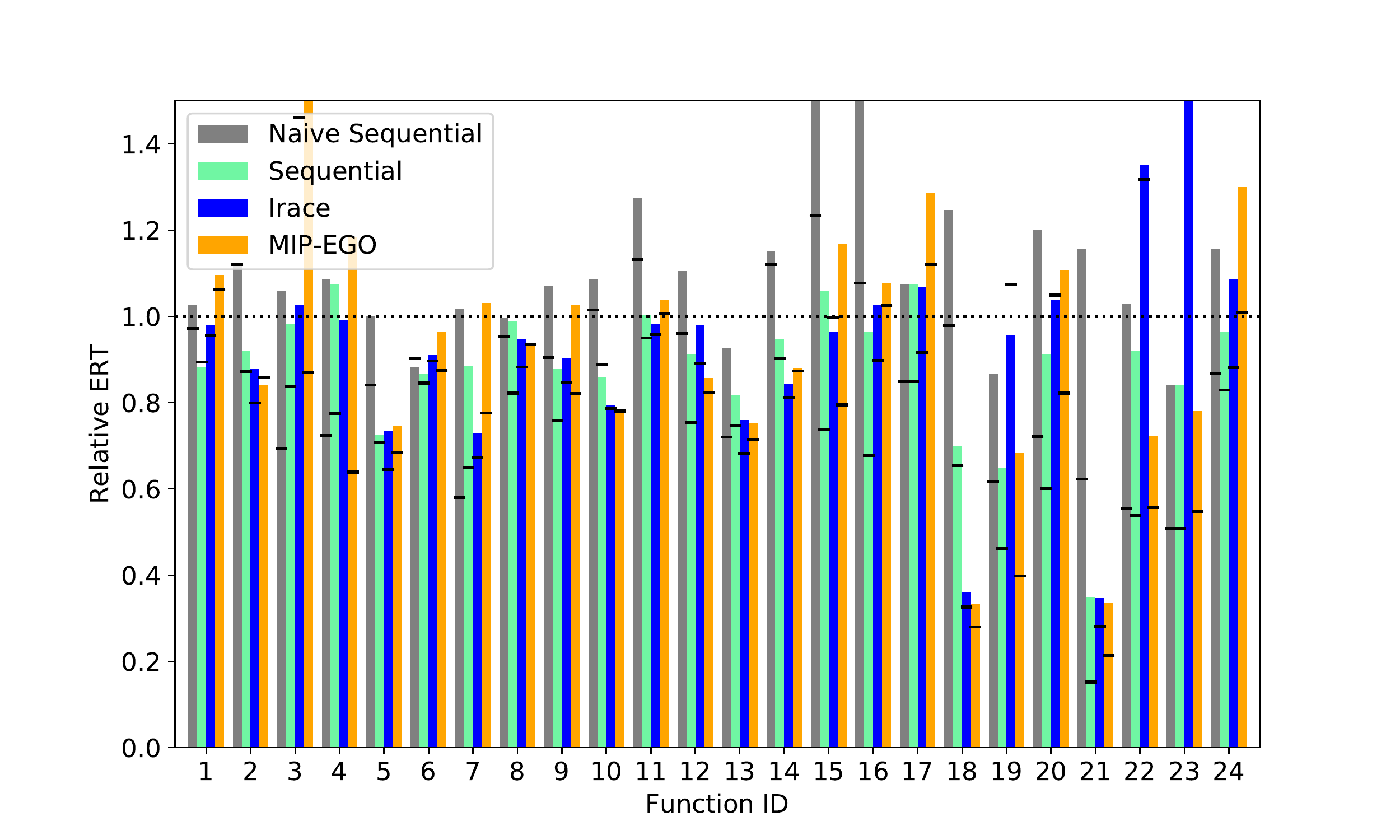}
    \caption{Relative ERTs against the best algorithm variant with default hyperparameters (targets chosen as in~\cite{research_project}) from running MIP-EGO and irace on the integrated selection and configuration space, as well as from the two sequential approaches. The `predicted' relative ERT (based on $25$ runs, with the exception of irace) is shown as a small black bar, whereas all other shown ERTs are based on $250$ verification runs. $y$-axis cut at 1.5 (full data set available at~\cite{data_thesis}). }
    %--- by the way: do you have this tuned data in your repository on online CMA-ES? Or did you create a new one for this part of your work? \died{This is a separate github page, I have added the reference here. I will update the page with the new figure generation code sometime next week when all figures have been finalized}} \carola{Does the color scheme work on a black-white printout? }}
    \label{fig:results_full}
    \vspace{-5pt}
\end{figure}

The results from running the integrated and sequential approaches on all 24 benchmark functions are shown in Figure~\ref{fig:results_full}. This figure shows that, in general, the ERT achieved by irace and MIP-EGO is comparable. Irace has a slight advantage, beating MIP-EGO on 14 out of 24 functions. However, both methods still manage to outperform the na\"ive sequential approach while using significantly fewer runs, and are only slightly worse than the more robust version of the sequential approach. As expected, all methods manage to outperform pure algorithm selection quite significantly.

% \begin{figure}
%     \centering
%     \includegraphics[width=0.5\textwidth, trim={60 10 50 20},clip]{img/integrated_vs_sequential_var2.pdf}
%     \caption{Resulting ERT (targets chosen as in~\cite{research_project}) from running MIP-EGO and irace on the integrated selection and configuration space,
%     as well as from the two sequential approaches. Configurations are selected based on $25$ runs for MIP-EGO (both the sequential and integrated variants) and for a variable number of runs for irace. The `predicted' ERT based on these runs is shown as a small black bar, whereas all other shown ERTs are based on $250$ verification runs. The dotted line represents the ERT of the best configuration with default hyperparameters.}
%     \label{fig:results_full}
% \end{figure}

\subsection{Comparison of MIP-EGO and Irace}
From the results presented in Figure~\ref{fig:results_full} it can be seen that the performance of the two integrated methods, MIP-EGO and irace, is quite similar. However, when introducing these methods, it was clear that their working principles differ significantly. To gain more understanding about how these results are achieved, three separate principles were studied: prediction error, balance between exploration and exploitation, and stability. 

\subsubsection{Prediction Error}\label{sec:pred_err}
%To investigate the differences between the two integrated methods, the prediction error can be studied in more detail. 
%From Figure~\ref{fig:results_full}, it seems to be the case 
The bars in Figure~\ref{fig:results_full} seem to indicate that the prediction error for irace is smaller than the one for MIP-EGO. This is indeed the case: the average prediction error is $10.6\%$ for irace, compared to $17.4\%$ for MIP-EGO, suggesting that the AHT values reported by irace are more robust than the ERTs given by MIP-EGO. However, we also note that there exist some outliers, for which the prediction error of irace is relatively large (up to $35\%$ for function 4). This happens because irace reports penalized AHT instead of ERT during the prediction-phase (see Definition~\ref{def:AHT}). However, these prediction errors for irace can be positive (i.e. overestimating the real ERT), whereas MIP-EGO always underestimates the actual ERT.

% \begin{figure}
%     \centering
%     \includegraphics[width=0.5\textwidth]{img/Prediction_error_onesearch_var2.pdf}
%     \caption{Prediction errors between ERT found during the run (penalized AHT in case of irace) and the ERT of $250$ runs for MIP-EGO, irace and sequential one-search-space methods, as well as the prediction error when selecting the best configuration using complete enumeration (and not running hyperparameter tuning).}
%     \label{fig:pred_errors}
% \end{figure}

\subsubsection{Exploration-Exploitation Balance}
While the prediction error is an important distinguishing factor between the two integrated methods, a much more important question to ask is how their search behaviour differs. This is best characterized by looking at the balance between exploration and exploitation, which we analyze by looking at the complete set of evaluated candidate (variant, hyperparameter)-pairs, and noting how many unique variants were explored after the initialization phase. For MIP-EGO, this number is on average $565$, while for irace it is only $112$. This leads us to conclude that MIP-EGO is very exploitative in the algorithm space, while irace is much more focused on exploitation of the hyperparameters.
% This metric is plotted in Figure~\ref{fig:exploration}, where it is clear that irace is very exploitative, while MIP-EGO is more focused on exploration of the configuration space. 
%When looking at the fraction of the candidates which only differ in terms of the continuous hyperparameters, so in other words how much of the budget was spent on tuning the hyperparameters of a single variant, this value is $78.6\%$. As opposed to only $2.6\%$ of them which contain unique variants. Even when including the initial random population, this only increases to $9.7\%$, while MIP-EGO achieves $77.8\%$ unique variants evaluated.
% \carola{is the following a correct interpretation of your data?}
On average, across all 24 benchmark functions, a fraction of $78.6\%$ of all candidates evaluated by irace differ only in terms of the continuous hyperparameters, whereas only $2.6\%$ of the evaluated (variant, hyperparameter) pairs contain unique variants. Even when including the initial random population, this value only increases to $9.7\%$, while MIP-EGO achieves an average fraction of $77.8\%$ unique variants evaluated.

% \begin{figure}
%     \centering
%     \includegraphics[width=0.5\textwidth]{img_2/Irace_mipego_number_confs_var2.pdf}
%     \caption{Number of distinct configurations explored after the initial sampling by both irace and MIP-EGO. MIP-EGO starts with a sample of $250$ points, while irace starts with $333$ candidates in the first race.}
%     \label{fig:exploration}
% \end{figure}

\begin{figure}
    \centering
    \includegraphics[width=0.5\textwidth, trim={25 5 20 32}, clip]{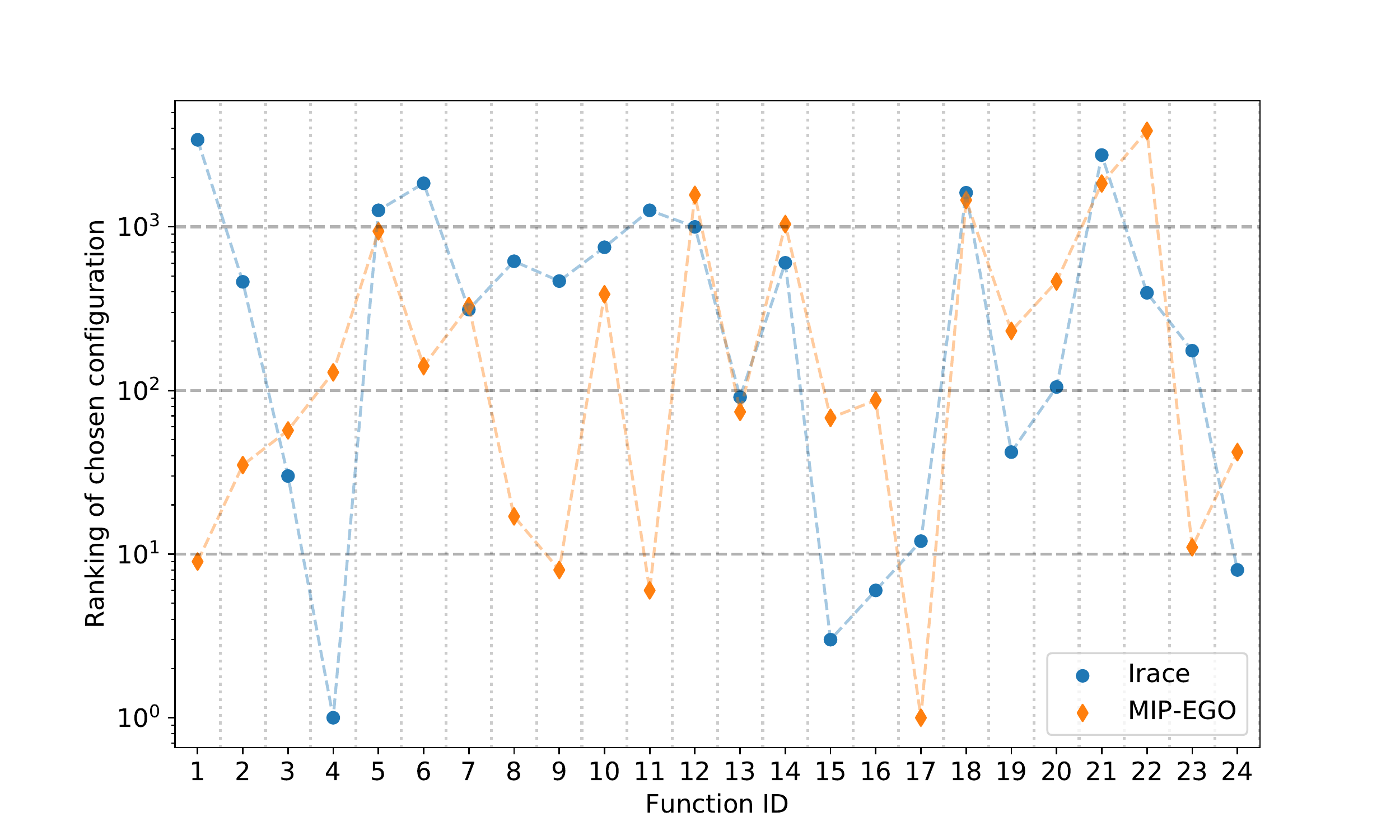}
    \caption{Original ranking (default hyperparameters) of the algorithm variant found by the integrated approaches. }
    % \carola{horizontal lines should be lighter, just like the vertical ones}}
    \label{fig:rank_dist}
    \vspace{-20pt}
\end{figure}

This difference in exploration-exploitation balance is expected to lead to a difference in variants found by irace and MIP-EGO, specifically in how these variants would rank with default hyperparameters.  This is visualized in Figure~\ref{fig:rank_dist}. From this figure, the differences between irace and MIP-EGO are quite clear. While MIP-EGO usually has better ranked variants, the median ranking is only $108$, as opposed to $428$ for irace. This confirms the findings of Section~\ref{sec:differences_impr}, where we saw there can be quite large differences in ranking before and after hyperparameter tuning. However, we still find that a larger focus on exploration yields a selection of variants which are ranked better on average.

\begin{figure}[!hb]
    \centering
    \includegraphics[width=0.5\textwidth, trim=0mm 0mm 0mm 15mm, clip]{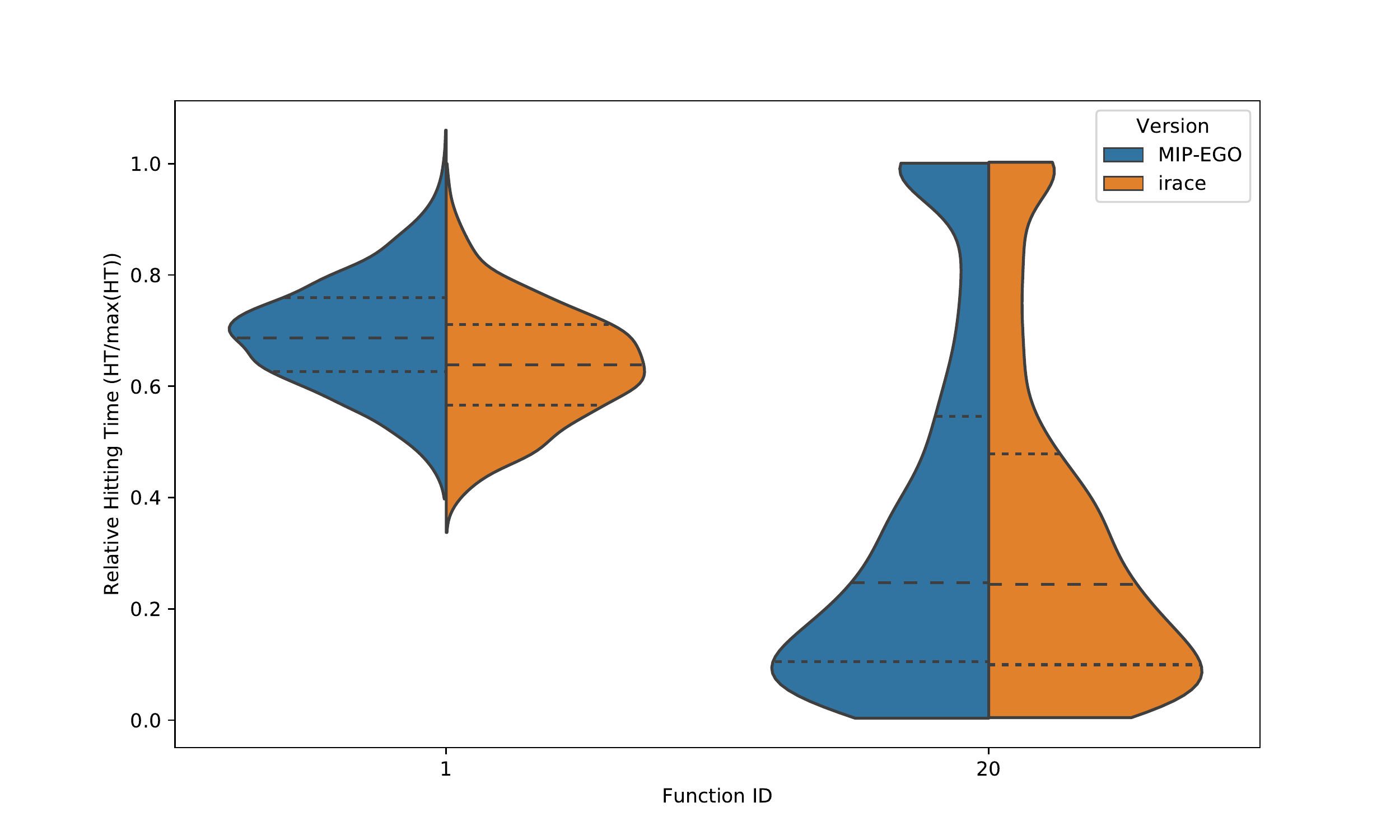}
    \caption{Distributions of relative hitting times (divided by the maximal hitting time observed for the function) of 15 (variant, hyperparameters)-pairs, resulting from 15 independent runs of the integrated approaches, each of which is run 250 times on the corresponding benchmark function.}
    \label{fig:violins_F20_F1}
    \vspace{-5pt}
\end{figure}

\subsubsection{Stability}
Finally, we study the variance in performance of the algorithm variants found by the two configurators. Since MIP-EGO is more exploitative, it might be more prone to variance than irace and thus less stable over multiple runs. To investigate this assumption, we select two benchmark functions and run both integrated methods 15 times. The resulting (variant, hyperparameters)-pairs are then rerun $250$ times, the runtime distributions of which are show in Figure~\ref{fig:violins_F20_F1}. For F20, there is a relatively small difference between irace and MIP-EGO, slightly favoring irace. This indicates that the exploitation done by irace is indeed beneficial, leading to slightly lower hitting times. For F1, this effect is much larger, since for F1 most variants behave quite similarly, so the more benefit can be gained by tuning the continuous hyperparameters relative to exploring the algorithm space.
% \begin{figure}
%     \centering
%     \includegraphics[width=0.5\textwidth]{img/Violins_F1.pdf}
%     \caption{Distributions of hitting times of 15 (configuration, hyperparameters)-pairs (resulting from 15 independent runs of the integrated approaches), each of which run 250 times on benchmark function F1.}
%     \label{fig:violins_F1}
% \end{figure}
% \begin{figure}
%     \centering
%     \includegraphics[width=0.5\textwidth]{img/Violins_F20.pdf}
%     \caption{Distributions of hitting times of 15 (configuration, hyperparameters)-pairs (resulting from 15 independent runs of the integrated approaches), each of which run 250 times on benchmark function F20.}
%     \label{fig:violins_F20}
% \end{figure}

\subsubsection{Summary}
A summary of the differences between the four methods studied in this paper can be seen in Table~\ref{tab:comp}. From  this,  we  can  see  that  the  differences  in  terms  of  performance between  the  integrated  and  sequential  methods  is  minimal,  while  they  require  a  significantly lower budget. This budget value is in no way optimized, so an even lower budget than the one used in our study might achieve similar results. This might especially be true for irace, since it uses most of its budget to evaluate very small changes in hyperparameter values. 
% \carola{add the figure from the slide deck? I would find that interesting} \died{I don't think that adds much here, as it is only about a single function with low sample size}

\begin{table}
\small

    \centering
    {\renewcommand{\arraystretch}{1.4}
        \begin{tabular}{
         @{}p{0.3\linewidth}|
            p{0.11\linewidth}
            p{0.11\linewidth}
            p{0.11\linewidth}
            p{0.11\linewidth}
            }
            \toprule
             & \textbf{Na\"ive\newline {Seq.}} & \textbf{Seq.} & \textbf{MIP-EGO} & \textbf{irace} \\
            \midrule
            Best on \# functions &  0 & 9 & 9 & 6 \\
            Avg. Impr. over best modular CMA-ES & $6.3\%$ & $24.7\%$ & $20.2\%$ & $20.7\%$ \\
            Avg. Impr. over default CMA-ES & $67.4\%$ & $73.0\%$ & $72.9\%$ & $72.5\%$\\
            Avg. Prediction Error &$23.2\%$ &$18.8\%$ & $17.4\%$ & $10.6\%$\\
            Budget (\# function evaluations)/1,000 & $\sim120$ %$120,\hspace{-1pt}200$ 
            & $150$ & $25$ & $25$\\
            $\%$ Unique CMA-ES variants explored & $95.8\%$ & $76.8\%$ & $77.8\%$ & $~9.7\%$\\
            \bottomrule
        \end{tabular}
        }
    \caption{Comparison of the four methods for determining (variant, hyperparameter)-pairs used in this paper. Seq.=sequential. Improvement over best modular CMA-ES refers to the relative improvement in ERT over the single best variant with default hyperparameters.}% Percentage of unique variants refers to how many of the evaluated candidates during the search contained unique variants.} %\carola{I abbreviated the budget. maybe you can now reduce the width?}}
    \label{tab:comp}
    \vspace{-10pt}
\end{table}
\section{Conclusions and Future Work}\label{sec:conclusion}

% \carola{The first part of this section can/should be drastically shortened}
% We have studied several ways of combining \textit{algorithm selection} and \textit{algorithm configuration} of modular CMA-ES variants into a single integrated approach. While a sequential execution of brute-force algorithm selection and hyperparameter tuning achieves decent improvements over pure algorithm selection, this improvement proved to be very unstable, caused by the large variance of the observed ERTs. This variance remains an issue, even when creating a more robust sequential approach which selects a larger set of algorithm variants for which to tune the hyperparameters. In addition, the sequential approaches require a large number of function evaluations, and quickly becomes prohibitive when new modules are added to the modEA framework. All these observations clearly illustrate a need for efficient and robust combined algorithm selection and configuration (CASH) methods. 

We have studied several ways of combining \textit{algorithm selection} and \textit{algorithm configuration} of modular CMA-ES variants into a single integrated approach. We have shown that a sequential execution of brute-force algorithm selection and hyperparameter is sub-optimal because the large variance present in the observed ERTs. In addition, the sequential approaches require a large number of function evaluations, and quickly becomes prohibitive when new modules are added to the modEA framework. This clearly illustrates a need for efficient and robust combined algorithm selection and configuration (CASH) methods. 

%It has been shown that a sequential execution of brute-force algorithm selection and hyperparameter tuning can achieve decent improvements over pure algorithm selection. However, this improvement proved to not be very stable, caused by the fact the inherent stochasticity added a large amount of variance to the observed ERTs. This variance remains an issue, even when creating a more robust sequential approach which selects a larger set of algorithm variants for which to tune the hyperparameters. Since it also holds that the impact of hyperparameter settings on performance is not constant over all functions and all variants, these sequential methods will never be able to guarantee optimality. The sequential approaches also require a large number of function evaluations, and this number grows exponentially as new modules are added to the modEA framework. All the above clearly illustrate a need for robust combined algorithm selection and configuration (CASH) methods. 

%For these integrated methods, we combine the hyperparameters tuned in the sequential approaches with the module activations in modEA to create one mixed-integer search space in order to tune both the algorithm variants and its hyperparameters at the same time. Both Irace and MIP-EGO are used to tune these (variant, hyperparameters)-pairs, which 
We have shown that both irace and MIP-EGO manage to solve the CASH problem for the modular CMA-ES. They outperform the results from the na\"ive sequential approach and show comparable performance to the more robust sequential method, and this at much smaller cost (up to a factor of $6$ in terms of function evaluations). %The integrated approach is therefore 
%While the results are similar, they are achieved using much less function evaluations, up to a factor of $6$, while being much more easily scalable to larger algorithm and hyperparameter spaces.

%As a final contribution, the differences between irace and MIP-EGO were studied in more detail. The differences in terms of performance between them were minimal, both for the hyperparameter tuning as for the integrated approaches. However, when looking closer at their internal processes, some significant differences in their working principles came to light. We have seen that MIP-EGO is more prone to underestimating ERTs, since it uses a static number of runs per candidate solution, whereas irace dynamically allocates its function evaluations based on the performance of the candidates. 
We have also observed that, for the integrated approach, MIP-EGO has a heavy focus on exploring the algorithm space, while irace spends most of its budget on tuning the continuous hyperparameters of a single variant. These differences were shown to lead to a slight benefit for irace on the sphere-function, but in general the difference in performance was minimal across the benchmark functions. This indicates that there is still room for improvement by combining the best parts from both methods into a single approach. This could take advantage of the dynamic allocation of runs to instances and adaptive capping which irace uses, as well as the efficient generation of new candidate solutions using the working principles of efficient global optimization, as done in MIP-EGO.

Another way to improve the viability of the integrated approaches studied in this paper would be to tune their maximum budget, as this was set arbitrarily in our experiments, and might be reduced significantly without leading to a large loss in performance.

We have focused in this work on the three hyperparameters selected in~\cite{BelkhirDSS17}. A straightforward extension of our work would be to consider the configuration of additional hyperparameters --- global ones that are present in all variants (such as the population size), but also conditional ones that appear only in some variants but not in others (i.e. the threshold value when the 'threshold convergence' module is turned on). While irace can deal with such conditional parameter spaces, MIP-EGO would have to be revised for this purpose. % \carola{is this correct?} \died{Yes, I think that is indeed the case}

% Another extension to this work is the adaptation of the proposed approaches to the configuration switching context. Configurations switching, as introduced in~\cite{van_rijn_ppns_2018_adpative}, has been shown to give promising results  in~\cite{research_project}. A brief summary of the configuration switching is included in Appendix~\ref{sec:switching}. To introduce hyperparameter tuning into the switching approach would mean to create a quintuple $(C_1,P_1,\sigma,C_2,P_2)$, where $(C_1, P_1)$ is the initial (configuration, hyperparameters)-pair, which switches to $(C_2, P_2)$ after target $\sigma$ (the splitpoint) is reached. To determine such a quintuple using the approach as described in~\cite{research_project} would be unfeasible, as the size of the search space grows too large to allow for complete enumeration. However, the selection of $(C_1, P_1)$ can be achieved using the integrated algorithm selection and configuration approaches described in this thesis. The main challenge for future work would then become the selection of the splitpoint and $(C_2, P_2)$.\bigskip

%As a final, long term extension, we propose to investigate the possibility to create an algorithm which can adapt its configuration and hyperparameters online, based on previous work done in~\cite{research_project}. 
Our long-term goal is to develop methods which adjust variant selection and configuration \emph{online}, i.e., while optimizing the problem at hand.
This could be achieved by building on exploratory landscape analysis~\cite{mersmann2011exploratory} and using a landscape-aware selection mechanism. Relevant features could be local landscape features such as provided by the flacco software~\cite{flacco} (this is the approach taken in~\cite{BelkhirDSS17}), but also the state of the CMA-ES-parameters themselves, and approach suggested in~\cite{PitraRH19}. % to determine whether a different algorithm design or a different configuration would be beneficial. 
We have analyzed the potential impact of such an online selection in~\cite{research_project}. %, and are next targeting to integrate the hyper-parameter optimization discussed in this work, as well as the mentioned landscape-aware selection.}  
Some initial work in determining how landscape features change during the search has been proposed in~\cite{jankovic2019adaptive}, but it was shown in~\cite{Renau2019features} that some of the local features provided by flacco are not very robust, so that a suitable selection of features is needed for the use in a landscape-aware algorithm design. 
%and exploration of which features are stable has been done in~\cite{Renau2019features}. Combining these works might eventually be able to lead to the creation of an adaptive, landscape-aware CMA-ES.

Finally, we are interested in generalizing the integrated algorithm selection and configuration approach studied in this work to more general search spaces, and in particular to possibly much more unstructured algorithm selection problems. For example, one could consider to extend the CASH approach to the whole set of algorithms available in the BBOB repository (some of these are summarized in~\cite{hansen2010comparing} but many more algorithms have been added in the last ten years since the writing of~\cite{hansen2010comparing}). Note that it is an open question, though, how well the here-studied configuration tools irace and MIP-EGO would perform on such an unstructured, categorical algorithm selection space. Note also that here again we need to take care of conditional parameter spaces, since the  algorithms in the BBOB data set have many different parameters that need to be set.   

\ack This work has been supported by the Paris Ile-de-France region, and by a public grant as part of the
Investissement d'avenir project, reference ANR-11-LABX-0056-LMH,
LabEx LMH.

\bibliography{bib_old}
\end{document}